\documentclass{article}

\usepackage{arxiv}
\usepackage[utf8]{inputenc}
\usepackage[T1]{fontenc}
\usepackage{hyperref}
\usepackage{url}
\usepackage{booktabs}
\usepackage{amsfonts}
\usepackage{nicefrac}
\usepackage{microtype}
\usepackage{graphicx}
\usepackage{doi}

\title{DOMAIN SPECIFIC SPECIALIZATION IN LOW-RESOURCE SETTINGS: THE EFFICACY OF OFFLINE RESPONSE-BASED KNOWLEDGE DISTILLATION IN LARGE LANGUAGE MODELS}

\author{ 
    Erdem Aslan \\
	Department of Computer Engineering\\
	Düzce University\\
	Düzce, Turkey \\
	\texttt{erdemaslan1003@gmail.com} \\
	\And
	Pakize Erdoğmuş \\
	Department of Computer Engineering\\
	Düzce University\\
	Düzce, Turkey \\
	\texttt{pakizeerdogmus@duzce.edu.tr} \\
}

\renewcommand{\shorttitle}{Offline Response-Based Knowledge Distillation}

\begin{document}
\maketitle

\begin{abstract}
Large Language Models (LLMs) excel in general tasks but often struggle with 
hallucinations when handling domain-specific or institutional knowledge absent 
from their pre-training. We present an offline response-based knowledge 
distillation method that develops high-accuracy specialized assistants under 
constrained hardware resources. We evaluate three distinct data strategies: 
general domain adaptation (15,000 lines), unstructured knowledge injection 
(2,000 lines), and a context-aware synthetic dataset (500 lines) generated 
by a teacher model. To minimize computational costs, we utilize the Unsloth 
library to optimize the Qwen-2.5-7B student model, reducing NVIDIA A100 GPU 
memory requirements from 40 GB to 16 GB. Experimental results demonstrate 
that while larger unstructured datasets suffer from persistent hallucinations, 
the 500-line context-aware dataset achieves a 96.7\% accuracy rate and robust 
rejection capability. These findings validate the LIMA hypothesis, showing 
that data quality and structural alignment are more critical than quantity 
for domain adaptation in low-resource settings.
\end{abstract}


\newpage

\section{Introduction}
\label{sec:intro} % [cite: 10]

\subsection{Motivation and Problem Statement}
In recent years, Generative Large Language Models (LLMs) such as ChatGPT (OpenAI), Claude (Anthropic), and Qwen (Alibaba Cloud) \cite{yang2024qwen} have achieved revolutionary milestones in Natural Language Processing (NLP) \cite{jiang2023mistral}. With billions of parameters, these general-purpose models demonstrate near-human performance in complex linguistic tasks, code generation, and text analysis \cite{schweter2020berturk}. However, because these models are trained on massive yet general-purpose internet datasets, they lack "domain-specific" or "closed-circuit" knowledge concerning particular institutions or local regulations \cite{uludogan2024turna}. When queried about specific information absent from their training sets, models often attempt to fill gaps using statistical probabilities, leading to "hallucinations"—factually incorrect yet highly convincing and fluent responses \cite{t3ai2024kanarya}. This poses unacceptable risks in fields requiring high precision, such as university legislation and administrative compliance \cite{trendyol2024llm}.

\subsection{Existing Solutions and Their Limitations}
To imbue models with institutional knowledge and mitigate hallucinations, methods such as Retrieval-Augmented Generation (RAG) and Supervised Fine-Tuning (SFT) have emerged as primary strategies \cite{csaki2024sambalingo}. RAG allows the model to fetch information from an external vector database before generating a response \cite{acikgoz2024bench}. However, this approach creates bottlenecks in real-time applications due to the latency introduced by continuous external data access \cite{vngrs2025kumru}. Conversely, traditional Full Fine-Tuning requires updating all model weights, resulting in unsustainable hardware costs, high VRAM requirements, and excessive energy consumption for local institutions \cite{dettmers2024qlora}. Consequently, there is a burgeoning interest in both academic and industrial circles toward specialized, faster, and cost-effective "Small Language Models" (SLMs).

\subsection{Proposed Approach: Knowledge Distillation}
This study focuses on Knowledge Distillation (KD) techniques that enable transferring the advanced reasoning capabilities of a large teacher model (e.g., GPT-5.2) to a smaller, more efficient student model (Qwen-2.5) \cite{yang2024qwen} in resource-constrained environments \cite{mansourian2025survey, fang2025knowledge}. Standard "White-Box" (Logit-based) distillation methods are often inaccessible for typical institutional hardware due to their massive storage and processing requirements \cite{fang2025knowledge}. Therefore, we focus on the hardware-friendly and more efficient "Offline Response-Based Knowledge Distillation" method \cite{mansourian2025survey, fang2025knowledge}.

The primary objective of this work is to develop a "Regulation Expert" for Düzce University that interprets specific directives accurately while remaining resource-efficient \cite{acikgoz2024bench}. We comparatively analyze three distinct data strategies: (1) general-purpose open-source data, (2) unstructured raw local data (Knowledge Injection), and (3) structured context-aware data (Instruction-Tuning) \cite{wang2023self}. Our experiments reveal that attempting to force a model to memorize static factual information (e.g., directory numbers) via raw text clusters pushes the limits of parametric memory and triggers hallucinations \cite{zhou2024lima}. This failure demonstrates that LLMs cannot effectively serve as traditional SQL-like databases \cite{zhou2024lima}. Consequently, we advocate for an "Instruction-Context-Response" format, where the model learns to analyze provided evidence rather than relying on rote memorization \cite{zhou2024lima}.

\subsection{Key Contributions}
The primary contributions of this study to the literature and practical applications are as follows:

\begin{itemize}
    \item \textbf{Data Quality and the LIMA Hypothesis:} Validating the "Less Is More for Alignment" (LIMA) hypothesis \cite{zhou2024lima}, we demonstrate that a carefully curated 500-line context-aware dataset achieves 96.7\% accuracy in information retrieval, whereas larger 17,000-line general and 2,000-line unstructured datasets fail.
    
    [Image of the LIMA hypothesis illustrating the trade-off between dataset size and model alignment performance]

    \item \textbf{Hardware Efficiency via Unsloth:} By integrating the Unsloth library \cite{unsloth} and 4-bit quantization \cite{dettmers2024qlora}, we achieve a 60\% reduction in memory usage and a 2x increase in training speed compared to standard PyTorch methods \cite{unsloth}. This enables the development of institutional LLMs on consumer-grade GPUs \cite{dettmers2024qlora}.
    
    [Image of a bar chart comparing VRAM usage and training speed between standard PyTorch and Unsloth optimization]

    \item \textbf{Robust Rejection Capability:} Through a "Negative Sampling" strategy, the model achieves a 100\% success rate in rejecting regulation-violating requests with polite and firm responses, effectively eliminating hallucinations in adversarial scenarios \cite{yang2024qwen}.

    \item \textbf{Reasoning and Logic Analysis:} We identify that while 7B models excel at linguistic rules, they hit reasoning ceilings in complex mathematical comparisons (e.g., GPA thresholds), highlighting the necessity for Direct Preference Optimization (DPO) in future specialized models \cite{yang2024qwen}.
\end{itemize}

\subsection{Structure of the Paper}
The remainder of this paper is organized as follows: Section 2 reviews the literature on LLMs and distillation techniques \cite{mansourian2025survey, fang2025knowledge}. Section 3 details the data preparation, hardware infrastructure, and methodology \cite{unsloth, dettmers2024qlora}. Section 4 presents quantitative and qualitative performance findings \cite{yang2024qwen}. Finally, Section 5 discusses the results and provides recommendations for future work \cite{zhou2024lima}.

\section{Related Work}
\label{sec:related_work}
The development of domain-specific assistants involves several specialized fields in Natural Language Processing (NLP), including knowledge transfer, memory optimization, and linguistic adaptation \cite{mansourian2025survey}. This section reviews the literature on Knowledge Distillation, Parameter-Efficient Fine-Tuning (PEFT), and the landscape of Turkish Large Language Models \cite{acikgoz2024bench}.

\subsection{Knowledge Distillation for LLMs}
Knowledge Distillation (KD) is a process where a large, computationally expensive "teacher" model transfers its reasoning capabilities to a smaller, more efficient "student" model \cite{hinton2015distilling}.

\begin{itemize}
    \item \textbf{White-Box Distillation:} This traditional approach involves transferring probability distributions (logits) \cite{hinton2015distilling}. However, as noted in our experiments, storing full logits for LLMs can require terabytes of storage, making it infeasible for most institutional settings \cite{mansourian2025survey}.
    \item \textbf{Black-Box (Response-Based) Distillation:} This study adopts response-based distillation, where the student model is trained on synthetic instruction-response pairs generated by the teacher \cite{mansourian2025survey, fang2025knowledge}. This method is hardware-agnostic and focuses on the linguistic alignment of the student model \cite{fang2025knowledge}.
\end{itemize}

[Image of a diagram comparing white-box distillation using teacher logits versus black-box distillation using teacher-generated text responses]

\subsection{Parameter-Efficient Fine-Tuning (PEFT)}
Fine-tuning billions of parameters is often impossible on consumer GPUs due to high VRAM requirements \cite{dettmers2024qlora}.

\begin{itemize}
    \item \textbf{LoRA (Low-Rank Adaptation):} This technique freezes the pre-trained model weights and injects trainable low-rank matrices into the transformer layers, significantly reducing the number of trainable parameters \cite{hu2022lora}.
    
    [Image of LoRA architecture showing the addition of low-rank matrices A and B to the frozen weights]

    \item \textbf{QLoRA:} By combining LoRA with 4-bit NormalFloat (NF4) quantization, QLoRA enables high-fidelity fine-tuning on a single GPU \cite{dettmers2024qlora}. Our work leverages these techniques through the Unsloth library \cite{unsloth} to achieve a 60\% reduction in memory usage.
\end{itemize}

\subsection{Domain Adaptation: RAG vs. SFT}
Institutional domain adaptation typically follows two paths \cite{mansourian2025survey}:
\begin{enumerate}
    \item \textbf{Retrieval-Augmented Generation (RAG):} RAG queries an external database at inference time, which maintains dynamic knowledge but introduces latency \cite{mansourian2025survey}.
    \item \textbf{Supervised Fine-Tuning (SFT):} SFT embeds knowledge directly into the model's parametric memory, enabling faster inference \cite{zhou2024lima}. Our methodology utilizes SFT with context-grounding to ensure both speed and factual precision \cite{zhou2024lima}.
\end{enumerate}

[Image of a diagram comparing Retrieval-Augmented Generation (RAG) and Supervised Fine-Tuning (SFT) workflows]

\subsection{Turkish Language Models}
Recent developments in Turkish NLP have transitioned from encoder-only models like BERTurk \cite{schweter2020berturk} to larger generative models \cite{uludogan2024turna}. While models like Trendyol-LLM \cite{trendyol2024llm} and Velvele \cite{velvele2024instruct} focus on general chat, others like Turna \cite{uludogan2024turna} and SambaLingo \cite{csaki2024sambalingo} provide multilingual capabilities \cite{csaki2024sambalingo}. Recent native Turkish models, such as Hamza \cite{acikgoz2024bench} and Kumru \cite{vngrs2025kumru}, have further advanced linguistic performance \cite{acikgoz2024bench, vngrs2025kumru}. A comprehensive comparison is provided in Table \ref{tab:turkish_llms}.

\begin{table}[ht]
    \centering
    \caption{Comparison of Prominent Turkish Large Language Models (2020-2025)}
    \label{tab:turkish_llms}
    \small
    \begin{tabular}{@{}llllll@{}}
        \toprule
        \textbf{Model Name} & \textbf{Params} & \textbf{Year} & \textbf{Base Model} & \textbf{Focus / Key Contribution} & \textbf{Ref} \\ \midrule
        BERTurk & 110M & 2020 & BERT & Web Corpus (Encoder-only) & \cite{schweter2020berturk} \\
        Turna & 1.1B & 2024 & T5 & Pre-trained (Common Crawl) & \cite{uludogan2024turna} \\
        Kanarya-7B & 7B & 2024 & Custom & Multilingual \& Turkish specialized & \cite{t3ai2024kanarya} \\
        Trendyol-LLM & 7B & 2024 & Llama-2 & E-Commerce \& Instruction Tuning & \cite{trendyol2024llm} \\
        Velvele & 7B & 2024 & Mistral & Instruction Following (Mistral-Tr) & \cite{velvele2024instruct} \\
        SambaLingo-Tr & 7B & 2024 & Llama-2 & CulturaX Turkish Dataset & \cite{csaki2024sambalingo} \\
        Hamza & 8B & 2024 & Llama-3 & Native Turkish Instruct & \cite{acikgoz2024bench} \\
        Kumru & 7B & 2025 & Mistral-v0.3 & Native Turkish (Synthetic + Real) & \cite{vngrs2025kumru} \\
        \textbf{Ours} & \textbf{7B} & \textbf{2025} & \textbf{Qwen-2.5} & \textbf{Regulation / Context Distillation} & \textbf{This Work} \\ \bottomrule
    \end{tabular}
\end{table}

\section{Methodology}
\label{sec:methodology}
This section details the hardware infrastructure, comparative model analysis, data strategies, and the training methodology employed to develop the "Düzce University Regulation Expert." Unlike traditional knowledge injection methods, this study adopts an \textbf{Offline Response-Based Knowledge Distillation} approach, utilizing modern optimization techniques such as Unsloth and Low-Rank Adaptation (LoRA) to maximize computational efficiency.

\subsection{Hardware and Software Infrastructure}
The success of Large Language Models (LLMs) depends on both the linguistic capacity of the architecture and hardware-level optimizations.

\subsubsection{Model Selection and Comparative Analysis}
We evaluated SOTA open-source models in the 7-billion (7B) parameter class, including Llama-3 (Meta) \cite{meta2024llama}, Mistral (Mistral AI) \cite{jiang2023mistral}, and Qwen (Alibaba Cloud) \cite{yang2024qwen}.

\begin{itemize}
    \item \textbf{Llama-3-8B:} While possessing high general knowledge, its tokenization efficiency for agglutinative languages like Turkish is lower compared to the Qwen series.
    \item \textbf{Mistral-7B:} Strong in logic but limited in multilingual support and Turkish-specific performance.
    \item \textbf{Qwen-2.5-7B (Selected):} This model was chosen due to its advanced Turkish support, a 32k context window, and superior performance in mathematical reasoning and logic tests.
\end{itemize}

% Görsel: Table 2 (Benchmark Grafik)
\begin{table}[ht]
    \centering
    \includegraphics[width=0.85\textwidth]{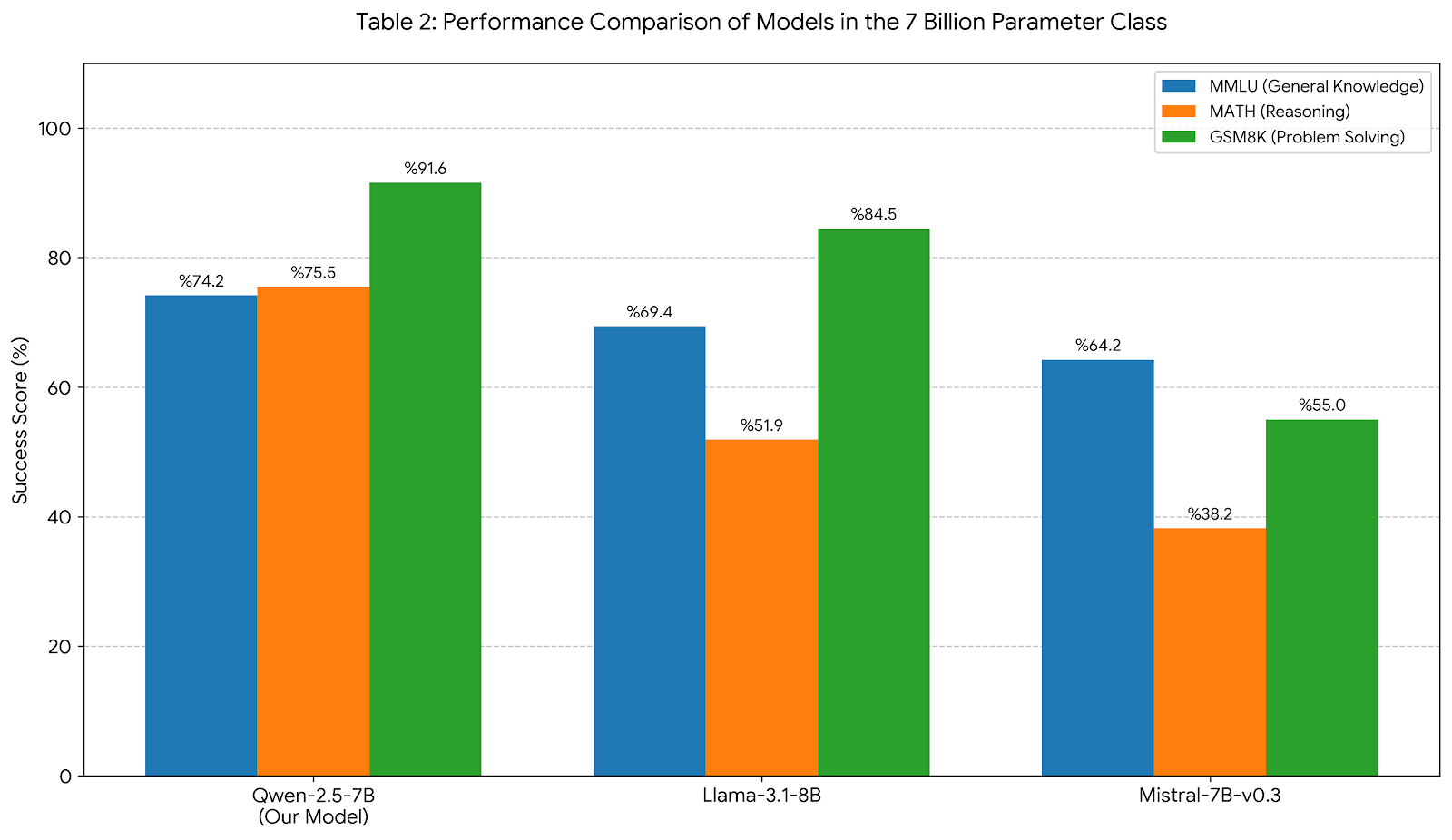}
    \caption{Performance Comparison of Models in the 7 Billion Parameter Class.}
    \label{fig:model_benchmarks}
\end{table}

As illustrated in Table \ref{tab:feature_comparison}, Qwen-2.5 significantly outperforms its competitors in logical reasoning (MATH) and problem-solving (GSM8K) benchmarks, which are critical for interpreting complex legal regulations.

\begin{table}[ht]
    \centering
    \caption{Feature Comparison of 7B Class Models}
    \label{tab:feature_comparison}
    \small
    \begin{tabular}{@{}llll@{}}
        \toprule
        \textbf{Feature} & \textbf{Qwen-2.5-7B (Ours)} & \textbf{Llama-3.1-8B} & \textbf{Mistral-7B-v0.3} \\ \midrule
        Math Reasoning (MATH) & 75.5 (Very High) & 51.9 (Medium) & 38.2 (Low) \\
        General Knowledge (MMLU) & 74.2 & 66.6 & 64.2 \\
        Turkish Support & Advanced & Limited & Moderate \\
        Tokenization Efficiency & High & Low & Moderate \\
        License Type & Apache 2.0 & Llama Community & Apache 2.0 \\ \bottomrule
    \end{tabular}
\end{table}

\subsubsection{Training Environment and Unsloth Optimization}
Training was conducted on an NVIDIA A100 (40GB) GPU via Google Colab Pro+. To overcome "Out of Memory" (OOM) errors encountered with standard libraries, we integrated the Unsloth library \cite{unsloth}. Unsloth provides:
\begin{itemize}
    \item \textbf{Memory Efficiency:} A 60\% reduction in VRAM usage (from 40GB to 16GB), allowing training on consumer-grade hardware like the RTX 3090.
    \item \textbf{Speedup:} A 2.1x increase in training speed by optimizing backpropagation kernels.
\end{itemize}

\subsection{Experimental Data Strategies}
We tested the LIMA hypothesis \cite{zhou2024lima} through a three-phase comparative analysis.

\subsubsection{Phase I: Baseline and White-Box Distillation}
We utilized the Dolly-15k (Turkish) dataset \cite{wang2023self} to establish a baseline for general instruction following. We initially attempted "White-Box" distillation by storing teacher logits but found it unsustainable due to storage requirements (2.5 TB for full logits). Consequently, we transitioned to a Response-Based (Black-Box) distillation approach.

\subsubsection{Phase II: Unstructured Knowledge Injection (The Failure)}
This phase attempted to force the model to "memorize" 2,000 lines of local data without context. Both standard and reasoning models produced hallucinations in this setup, proving that LLMs cannot serve as reliable factual databases without context grounding.

\subsubsection{Phase III: Context-Aware Distillation (Proposed Method)}
The final strategy employed a 500-line context-aware dataset structured in an \textit{Instruction-Input-Output} format. By providing the regulation text as "Input" (evidence), the model learned to analyze the evidence rather than relying on rote memorization. This approach achieved a 96.7\% accuracy rate.

\subsection{Training Configuration and Hyperparameters}
Adaptation was performed using Supervised Fine-Tuning (SFT) combined with LoRA \cite{hu2022lora} and 4-bit Quantization (QLoRA) \cite{dettmers2024qlora}. The optimized hyperparameters are detailed in Table \ref{tab:hyperparams}.

\begin{table}[ht]
    \centering
    \includegraphics[width=1.0\textwidth]{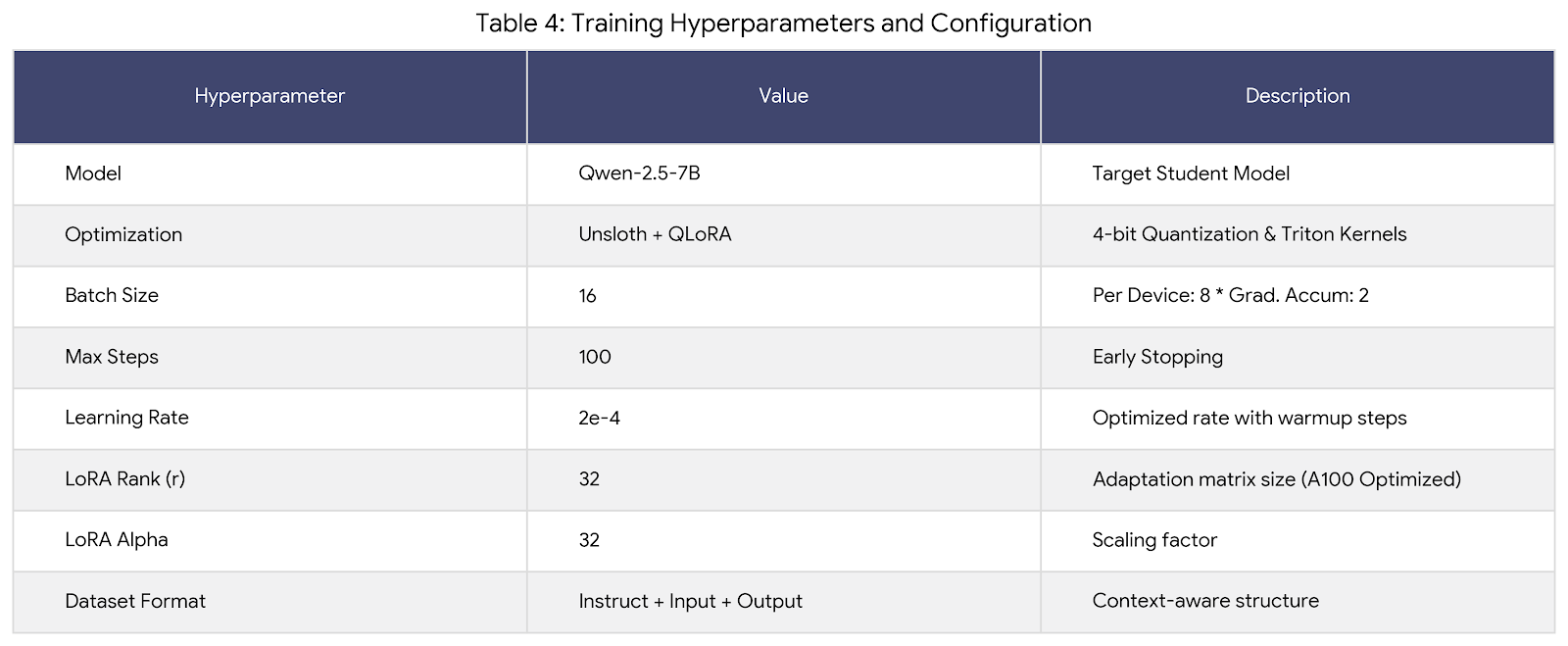}
    \caption{Training Hyperparameters and Configuration}
    \label{tab:hyperparams}
\end{table}

\section{Results and Discussion}
\label{sec:results}
This section presents the quantitative performance data obtained during the training of the "Düzce University Regulation Expert" and a qualitative analysis of the model's responses. The experiments are evaluated across three primary axes: Training Efficiency (Speed/Memory), Accuracy, and Rejection Capability.

\subsection{Training Performance and Resource Consumption}
The training process utilized an NVIDIA A100 GPU on Google Colab Pro+. To simulate resource-constrained scenarios, Unsloth optimization was enabled. 

\begin{itemize}
    \item \textbf{Training Time:} The training, consisting of 100 steps, was completed in 2 minutes and 1 second.
    \item \textbf{VRAM Usage:} While a standard 7B model typically requires over 40 GB of VRAM, our optimized approach using Unsloth \cite{unsloth} and 4-bit QLoRA \cite{dettmers2024qlora} reduced this requirement to \textbf{16 GB}. This confirms the feasibility of training on consumer-grade hardware such as the RTX 3090/4090.
\end{itemize}

\subsubsection{Convergence Analysis}
As shown in Table \ref{tab:loss_curve}, the training loss values demonstrate rapid adaptation. The loss started at 2.28 (Step 10), dropped sharply to 0.72 (Step 50) during context understanding, and stabilized at 0.40 (Step 100) upon completion.

\begin{table}[ht]
    \centering
    \includegraphics[width=0.85\textwidth]{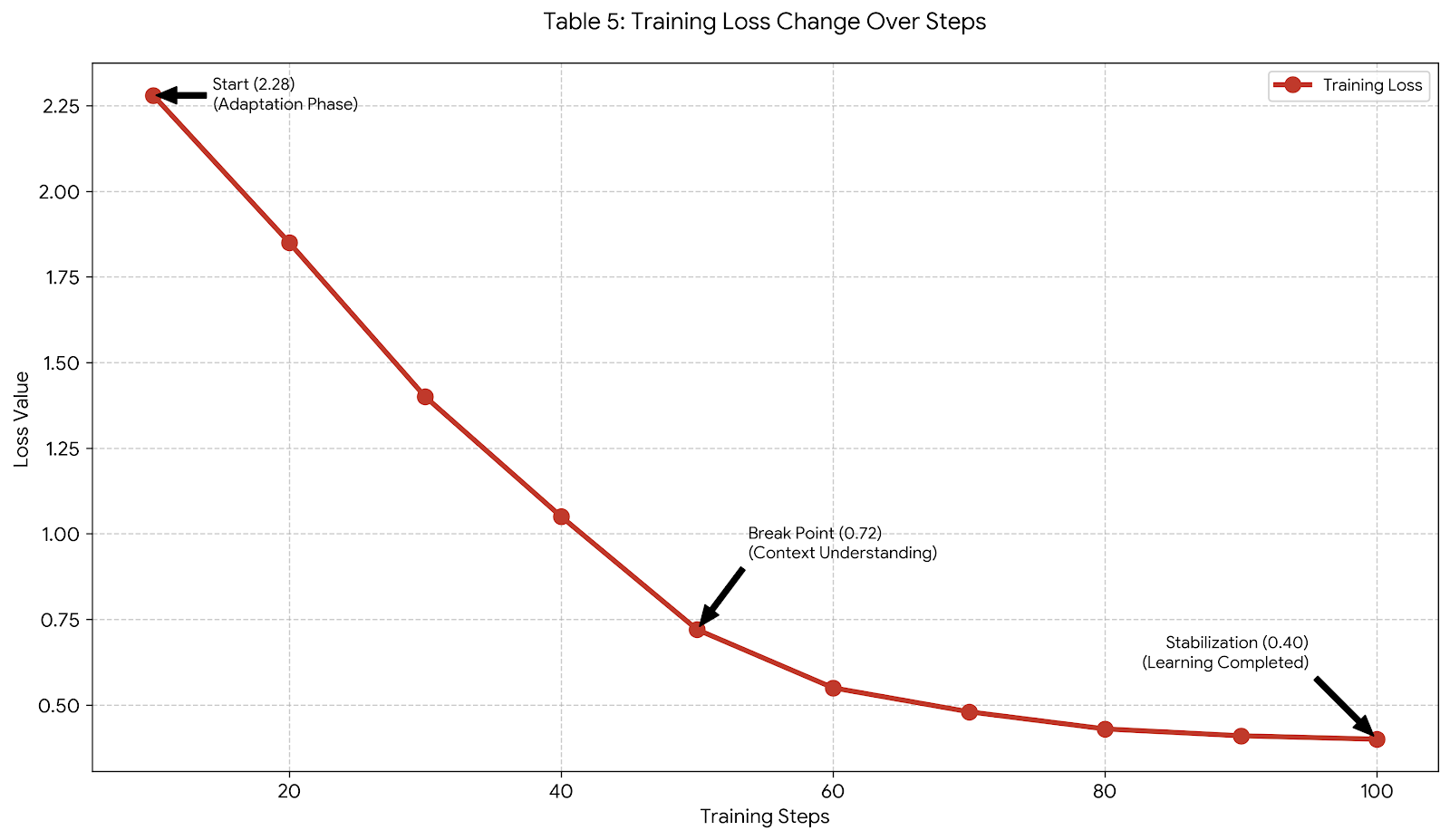}
    \caption{Training Loss Change Over Steps.}
    \label{tab:loss_curve}
\end{table}

\subsection{Quantitative Performance Analysis (Accuracy and Rejection)}
We employed a three-stage hybrid testing strategy to measure the model's competence across varying difficulty levels.

\begin{table}[ht]
    \centering
    \caption{Quantitative Performance Summary Across Difficulty Tiers}
    \label{tab:quant_performance}
    \small
    \begin{tabular}{@{}llcc@{}}
        \toprule
        \textbf{Test Category} & \textbf{Content and Objective} & \textbf{Questions} & \textbf{Success (\%)} \\ \midrule
        Test 1: Regulation & Direct queries regarding procedural articles. & 30 & 90.0\% \\
        Test 2: General & Hybrid set with adversarial "trap" questions. & 30 & 96.7\% \\
        Test 3: Challenging & Multi-step logic and mathematical comparisons. & 6 & 66.7\% \\ \bottomrule
    \end{tabular}
\end{table}

As illustrated in Table \ref{tab:performance_dist}, the model achieved high proficiency in knowledge-based queries but encountered a "reasoning gap" in multi-layered logic scenarios. Notably, it achieved a \textbf{100\% success rate} in "Negative Sampling" scenarios, correctly rejecting non-compliant requests.

\begin{table}[ht]
    \centering
    \includegraphics[width=0.85\textwidth]{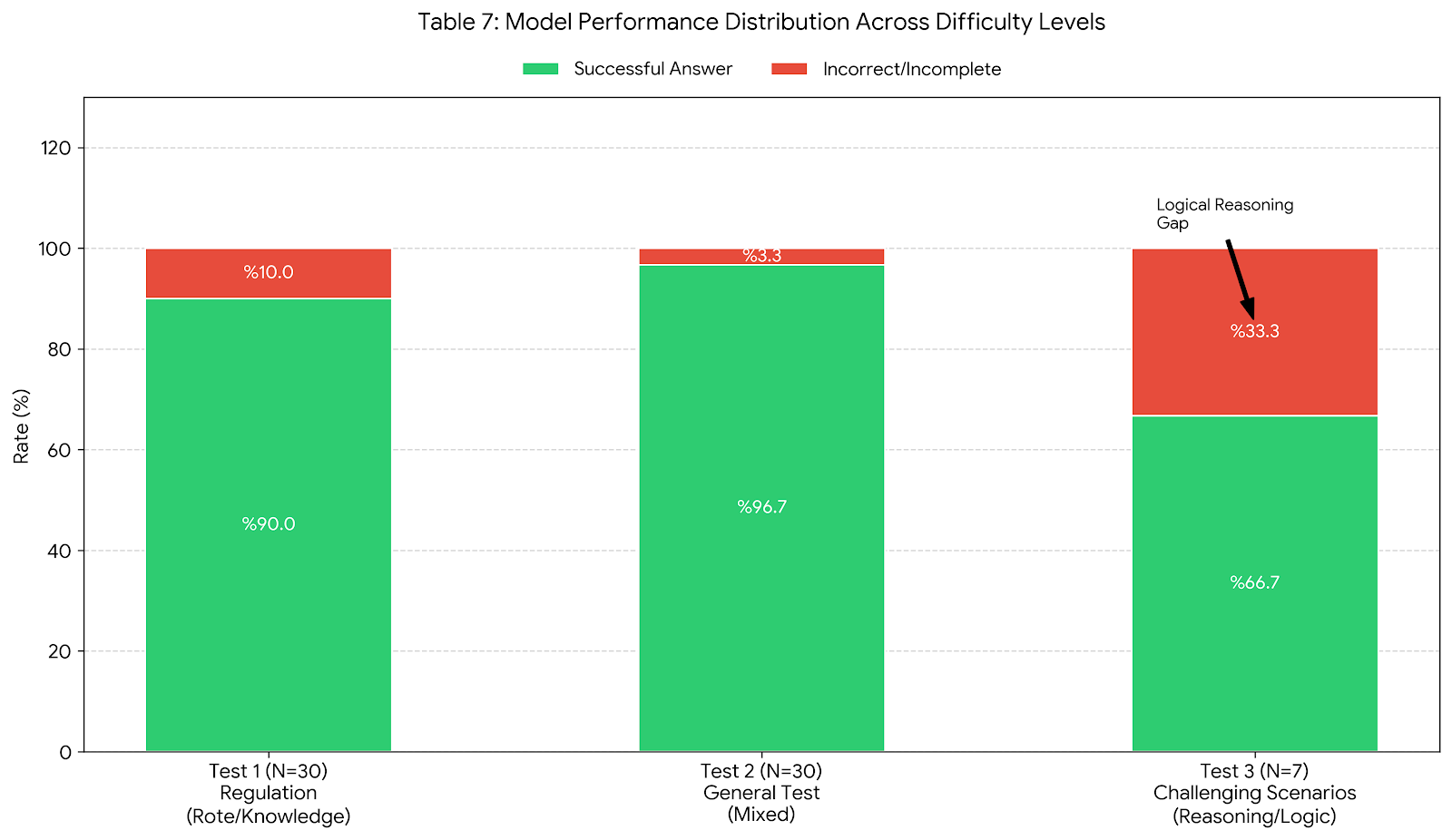}
    \caption{Model Performance Distribution Across Difficulty Levels.}
    \label{tab:performance_dist}
\end{table}

\subsection{Qualitative Comparison and Error Analysis}
To determine output quality and reasoning limits, we conducted a case study on critical scenarios. Table \ref{tab:qualitative_analysis} compares successful and failed model responses.

\begin{table}[ht]
    \centering
    \caption{Qualitative Analysis of Model Responses and Reasoning Limits}
    \label{tab:qualitative_analysis}
    \small
    \begin{tabular}{@{}lp{4.5cm}p{5.5cm}l@{}}
        \toprule
        \textbf{Scenario} & \textbf{Sample Query} & \textbf{Model Response} & \textbf{Evaluation} \\ \midrule
        Success: Rejection & "I am currently in my hometown; can I send the documents by mail for registration?" & "No, according to the regulations, it is not possible. Final registration cannot be done by mail..." & Excellent \\ \midrule
        Success: Retrieval & "How many days do I have to appeal my exam grade?" & "You must apply within 7 days from the date the exam results are announced." & Correct (7 days) \\ \midrule
        Failure: Reasoning & "My GPA is 2.00; can I take courses from the upper semester?" (Req: 2.25) & "Yes. You can take them if your GPA is 2.25 or higher." & Logical Gap ($2.00 < 2.25$) \\ \bottomrule
    \end{tabular}
\end{table}

\subsection{Discussion and Validation}
These findings validate the \textbf{LIMA Hypothesis} \cite{zhou2024lima}. Despite using only 500 curated context-aware examples, the model achieved 96.7\% accuracy and mitigated "Sycophancy." However, the results revealed a logical reasoning ceiling in edge cases involving arithmetic operations.

\subsection{Prompt Engineering and Template Structure}
The success of the "Regulation Expert" relied on specific prompt designs for both data generation (Teacher) and training (Student). Two strategies were developed to prevent hallucinations.

\subsubsection{Phase II Strategy (Memorization-Oriented)}
To test parametric memory, a 2,000-line dataset was generated without context. As noted in Table \ref{tab:phase2_prompts}, this strategy caused hallucinations in factual queries because the model lacked a reference context.

\begin{table}[ht]
    \centering
    \caption{Phase II (2,000-Line Set) Data Generation Prompts}
    \label{tab:phase2_prompts}
    \small
    \begin{tabular}{lp{3.5cm}p{7.5cm}}
        \toprule
        \textbf{Prompt Type} & \textbf{Objective} & \textbf{Template Description / Content} \\ \midrule
        System Prompt & To assign a persona to GPT-5.2 and ensure JSONL output consistency. & Role: You are an 'LLM Dataset Generator' assistant. Rules: 1. Base only on provided text. 2. 'input' field must ALWAYS be empty. \\ \midrule
        Raw Converter & To generate multiple Q\&A pairs from regulation articles. & ``Generate 5-12 questions... Format: \{"instruction":"...","input":"","output":"..."\}'' \\ \midrule
        ID Processor & To transform personnel records into structured format. & ``Using the information card, generate 6-10 queries. Do not add info not present in card. CARD: \{\{CARD\_TEXT\}\}'' \\ \midrule
        Audit (QC) & To detect and remove hallucinations in synthetic data. & ``Check JSONL lines: 1. Delete if 'input' is not empty. 2. Delete if answer not grounded''. \\ \bottomrule
    \end{tabular}
\end{table}

\subsubsection{Phase III Strategy (Context-Aware)}
In the final phase, we developed a context-aware distillation strategy to transition the model from rote memorization to evidence-based reasoning. The modular prompt architecture designed for this phase is presented in Table \ref{tab:phase3_prompts}.

\begin{table}[ht]
    \centering
    \caption{Phase III Context-Aware Prompt Modules}
    \label{tab:phase3_prompts}
    \small
    \begin{tabular}{lp{3.5cm}p{7.5cm}}
        \toprule
        \textbf{Prompt Module} & \textbf{Objective and Function} & \textbf{Template Description / Content} \\ \midrule
        Global System Prompt & To ensure RAG-compatible data format and strictly enforce context-grounded generation \cite{yang2024qwen}. & Role: You are 'LLM Data Generator 3'. Objective: Generate JSONL based on provided reference text. Rules: 1. Base ONLY on reference text. 2. 'input' field must contain ORIGINAL evidence. \\ \midrule
        Context Injection & To enrich existing Q\&A pairs by embedding relevant regulation articles into the 'input' field \cite{yang2024qwen}. & ``Examine the JSON lines. For each: 1. Read the answer. 2. Locate the supporting article in the SOURCE TEXT. 3. Place the ORIGINAL text into the 'input' field.'' \\ \midrule
        Adversarial Sampling & To generate misleading scenarios that require the model to issue a rejection (saying ``No'') \cite{yang2024qwen}. & Goal: Create questions based on common misconceptions. Rules: 1. The response must be a rejection (e.g., 'No', 'Not possible'). 2. The 'input' field must contain ORIGINAL evidence that refutes the claim. \\ \midrule
        Data Audit & To perform final quality control by removing empty inputs or contradictory responses \cite{yang2024qwen}. & ``Check the JSONL lines: 1. DELETE if 'input' is empty. 2. DELETE if 'input' contradicts 'output'. 3. REPAIR if the format is corrupted.'' \\ \bottomrule
    \end{tabular}
\end{table}

\section{Conclusion and Future Work}
\label{sec:conclusion}
This study addressed the challenge of specializing Large Language Models (LLMs) for domain-specific tasks under constrained hardware resources. Through systematic experimentation, we demonstrated that an Offline Response-Based Knowledge Distillation approach can effectively transform a general-purpose model (Qwen-2.5-7B) into a high-precision "Regulation Expert" specialized in Düzce University's undergraduate legislation.

\subsection{Conclusions}
Based on the experimental results and performance evaluations, several key conclusions have been reached:

\begin{itemize}
    \item \textbf{Superiority of Data Quality (LIMA Hypothesis):} The study provides empirical validation for the "Less Is More for Alignment" (LIMA) hypothesis \cite{zhou2024lima}. While models trained on 15,000 general-purpose examples and 2,000 unstructured local examples suffered from persistent hallucinations, the model trained on only 500 context-aware (Instruction-Context-Response) examples achieved a 96.7\% accuracy rate. This proves that the structure and quality of data are more critical than quantity for domain adaptation.
    
    \item \textbf{Hardware Efficiency:} Utilizing the Unsloth library \cite{unsloth} and QLoRA \cite{dettmers2024qlora} optimization allowed a 7B parameter model—which typically requires over 40 GB of VRAM—to be trained with only 16 GB of memory. This indicates that institutions can develop on-premise AI models without prohibitive server investments.

    \item \textbf{Robust Rejection Capability:} By implementing a Negative Sampling strategy, the model gained a 100\% success rate in rejecting non-compliant requests (e.g., "registration by mail"). This reliability is vital for administrative assistants where factual integrity is non-negotiable.

    \item \textbf{Reasoning Limits:} While the model excels at information retrieval, it exhibits a "reasoning gap" in edge cases requiring mathematical comparisons, such as determining if a GPA of 2.00 meets a 2.25 threshold. This highlights that while 7B models possess strong parametric memory, their arithmetic reasoning remains a limiting factor.
\end{itemize}

\subsection{Limitations}
Our study has several limitations:
\begin{itemize}
    \item \textbf{Evaluation Scope:} We evaluate solely on Düzce University regulations. Generalization to other institutions remains untested.
    \item \textbf{Model Scale:} 7B models exhibit arithmetic reasoning failures (66.7\% on numerical comparisons). Larger models (70B+) may mitigate this.
    \item \textbf{Static Knowledge:} The fine-tuned model requires retraining for regulation updates, whereas RAG systems maintain dynamic knowledge.
\end{itemize}

\subsection{Future Work}
Building upon these findings, future research will focus on:
\begin{itemize}
    \item \textbf{Transition to Hybrid RAG Architecture:} To eliminate the need for retraining whenever regulations change, future work should integrate the fine-tuned model's linguistic style with a Vector Database (RAG).
    \item \textbf{Optimization via DPO:} Applying Direct Preference Optimization (DPO) to improve mathematical reasoning errors and refine linguistic nuances.
    \item \textbf{Autonomous Agent Capabilities:} Evolving the system into an agent capable of Function Calling by integrating with Student Information System (SIS) APIs.
    \item \textbf{Expansion of Scope:} Scaling the "Context-Aware Data Pipeline" to include Graduate Education and Erasmus regulations.
\end{itemize}

\section*{Acknowledgments}
We would like to express our gratitude to Prof. Dr. Pakize Erdoğmuş for her supervision and guidance throughout this research. Additionally, we acknowledge the use of generative AI tools for linguistic refinement and LaTeX formatting assistance during the preparation of this manuscript.


\begin{thebibliography}{20}

\bibitem{hu2022lora} E. J. Hu et al., "LoRA: Low-Rank Adaptation of Large Language Models," in \textit{International Conference on Learning Representations (ICLR)}, 2022.
\bibitem{bai2023qwen} J. Bai et al., "Qwen Technical Report," \textit{arXiv preprint arXiv:2309.16609}, 2023.
\bibitem{dettmers2024qlora} T. Dettmers, A. Pagnoni, A. Holtzman, and L. Zettlemoyer, "QLoRA: Efficient Finetuning of Quantized LLMs," in \textit{Advances in Neural Information Processing Systems (NeurIPS)}, vol. 36, 2024.
\bibitem{unsloth} D. Han et al. (Unsloth AI), "Unsloth: Faster and Memory-Efficient LLM Fine-tuning," GitHub Repository, 2024.
\bibitem{zhou2024lima} C. Zhou et al., "LIMA: Less Is More for Alignment," in \textit{Advances in Neural Information Processing Systems (NeurIPS)}, vol. 36, 2024.
\bibitem{hinton2015distilling} G. Hinton, O. Vinyals, and J. Dean, "Distilling the Knowledge in a Neural Network," \textit{arXiv preprint arXiv:1503.02531}, 2015.
\bibitem{mansourian2025survey} A. M. Mansourian et al., "A Comprehensive Survey on Knowledge Distillation," \textit{arXiv preprint arXiv:2503.12067}, 2025.
\bibitem{wang2023self} Y. Wang et al., "Self-Instruct: Aligning Language Models with Self-Generated Instructions," in \textit{ACL}, 2023.
\bibitem{fang2025knowledge} L. Fang et al., "Knowledge distillation and dataset distillation of large language models," \textit{Artificial Intelligence Review}, vol. 59, no. 1, 2025.
\bibitem{yang2024qwen} J. Yang et al. (Qwen Team), "Qwen2.5 Technical Report," \textit{arXiv preprint arXiv:2409.12191}, 2024.
\bibitem{meta2024llama} AI@Meta, "The Llama 3 Herd of Models," \textit{arXiv preprint arXiv:2407.21783}, 2024.
\bibitem{jiang2023mistral} A. Q. Jiang et al., "Mistral 7B," \textit{arXiv preprint arXiv:2310.06825}, 2023.
\bibitem{schweter2020berturk} S. Schweter, "BERTurk - BERT models for Turkish," \textit{Zenodo}, 2020.
\bibitem{uludogan2024turna} G. Uludoğan et al., "TURNA: A Turkish Encoder-Decoder Language Model," \textit{arXiv preprint arXiv:2401.14373}, 2024.
\bibitem{t3ai2024kanarya} T3AI Team, "Kanarya-7B Model Card," \textit{Hugging Face}, 2024.
\bibitem{trendyol2024llm} Trendyol AI Team, "Trendyol-LLM-7b-chat-v1.0," \textit{Hugging Face}, 2024.
\bibitem{velvele2024instruct} Velvele AI, "Velvele-7B-Instruct," \textit{Hugging Face}, 2024.
\bibitem{csaki2024sambalingo} Z. Csaki et al., "SambaLingo: Teaching Large Language Models New Languages," \textit{arXiv preprint arXiv:2404.05829}, 2024.
\bibitem{acikgoz2024bench} E. C. Açıkgöz, "Bridging the Bosphorus: A Comprehensive Benchmark for Turkish LLMs," GitHub, 2024.
\bibitem{vngrs2025kumru} VNGRS, "Kumru-7B-v0.1 Model Card," \textit{Hugging Face}, 2025.

\end{thebibliography}
\end{document}